\documentclass[iicol]{sn-jnl}

\usepackage{graphicx}
\usepackage[margin=10pt]{subfig}
\usepackage{xcolor}
\usepackage{placeins}
\usepackage{booktabs}
\usepackage{amsmath}
\usepackage{siunitx}
\usepackage{tikz}
\usetikzlibrary{positioning}
\usepackage{url}

\usepackage{breakurl}
\begin{document}

\title[Less can be more: representational vs. stereotypical gender bias in facial expression recognition]{Less can be more: representational vs. stereotypical gender bias in facial expression recognition}

\author*[1]{\fnm{Iris} \sur{Dominguez-Catena}}\email{iris.dominguez@unavarra.es}

\author[1]{\fnm{Daniel} \sur{Paternain}}\email{daniel.paternain@unavarra.es}

\author[1]{\fnm{Aranzazu} \sur{Jurio}}\email{aranzazu.jurio@unavarra.es}

\author[1]{\fnm{Mikel} \sur{Galar}}\email{mikel.galar@unavarra.es}

\affil*[1]{\orgdiv{Department of Statistics, Computer Science and Mathematics}, \orgname{Public University of Navarre (UPNA)}, \orgaddress{\street{Arrosadia Campus}, \city{Pamplona}, \postcode{31006}, \state{Navarre}, \country{Spain}}}

\abstract{

Machine learning models can inherit biases from their training data, leading to discriminatory or inaccurate predictions. This is particularly concerning with the increasing use of large, unsupervised datasets for training foundational models. Traditionally, demographic biases within these datasets have not been well-understood, limiting our ability to understand how they propagate to the models themselves. To address this issue, this paper investigates the propagation of demographic biases from datasets into machine learning models. 
We focus on the gender demographic component, analyzing two types of bias: representational and stereotypical. For our analysis, we consider the domain of facial expression recognition (FER), a field known to exhibit biases in most popular datasets. We use Affectnet, one of the largest FER datasets, as our baseline for carefully designing and generating subsets that incorporate varying strengths of both representational and stereotypical bias. Subsequently, we train several models on these biased subsets, evaluating their performance on a common test set to assess the propagation of bias into the models' predictions.
Our results show that representational bias has a weaker impact than expected. Models exhibit a good generalization ability even in the absence of one gender in the training dataset. Conversely, stereotypical bias has a significantly stronger impact, primarily concentrated on the biased class, although it can also influence predictions for unbiased classes. 
These results highlight the need for a bias analysis that differentiates between types of bias, which is crucial for the development of effective bias mitigation strategies.

}


\maketitle

\section{Introduction}\label{sec:intro}

    In the last decades, advancements in Machine Learning algorithms have granted the field a rapid growth. The robustness of these systems and, in particular, their ability to generalize effectively when encountering novel data, has allowed them to make use of the vast amounts of information available online~\cite{Jordan2015}. This data is often consolidated in the form of large public datasets, supporting independent data gathering efforts and model development~\cite{Kreutzer2022}, and providing unified benchmarks for method comparison~\cite{Zhu2022}.
    
    As datasets have grown in size over time, there has been a notable shift from meticulously curated data gathered under controlled laboratory settings to data abundantly sourced from the wild, colloquially termed In The Wild (ITW) data~\cite{Prabhu2020}. This transition has supported the development of advanced deep learning models, specialized in extracting any shared complex patterns. Thanks to it, substantial breakthroughs have occurred, especially in complex tasks such as natural language processing~\cite{Brown2020} and computer vision~\cite{Maaz2023}.
    
    Unfortunately, ITW-gathered datasets, while cheaper to acquire, often contain unwanted patterns and properties, known as biases~\cite{Mehrabi2021,Birhane2021}. These biases, if left unaddressed, have the potential to propagate into machine learning models during training~\cite{Suresh2021}. Models contaminated by these biases may behave unreliably and unpredictably. When these patterns relate to protected demographic attributes, such as race, gender, age, or religion, the models have been shown to replicate and exacerbate social discrimination: making inaccurate predictions for minority groups~\cite{Buolamwini2018}, perpetuating stereotypes~\cite{Garrido-Munoz2023} and even directly discriminating users~\cite{Adams-Prassl2023}, among others.
    
    One of the problems that experimented this change to ITW datasets is Facial Expression Recognition (FER)~\cite{Nidhi2023}. FER is a computer-vision-based branch of affective computing that focuses on automatically detecting facial expressions from images. These systems have a wide range of applications, including driver drowsiness detection~\cite{Mou2023}, assistive robotics~\cite{Nimmagadda2022}, healthcare~\cite{Werner2022} and general human-computer interaction~\cite{Bakariya2023}. By accurately recognizing facial expressions, FER technology can enhance user experiences, improve human-computer interaction, and facilitate emotion-aware applications in various domains.
    
    Despite the popularity of the FER approach, demographic bias problems have already been found in various FER systems, both in research models~\cite{Xu2020,Domnich2021,Deuschel2021,Jannat2021,Poyiadzi2021} and in commercial systems~\cite{Kim2021,Ahmad2022}. These biases are often studied in model predictions, where harm could be done to a potential user. Nonetheless, the origin of the model bias has been linked to the training data~\cite{Deuschel2021,Xu2020,Dominguez-Catena2022}, which has supported the specific study of bias at the dataset level~\cite{Dominguez-Catena2024}. In this study, two key types of bias commonly present in datasets were identified, namely representational and stereotypical bias. Representational bias, in this context, refers to the general presence or absence of certain demographic groups in a dataset as a whole. Conversely, stereotypical bias refers to a relative over or underrepresentation of certain demographic groups in one of the target classes. Although metrics have been proposed to quantify both types of bias, further research is still needed to understand how and in what amount these biases transfer to the trained model. 
    

    Thus, the objective of this work is to analyze the propagation of demographic bias from the datasets into the models and, in particular, we aim to observe if different effects can be attributed to representational and stereotypical biases. We focus on FER as our case study, taking advantage of previous research on the bias properties of its datasets~\cite{Dominguez-Catena2024}. Additionally, we focus on the apparent gender of the subjects for this study. This demographic property can be approximated by a binary variable, which facilitates both the definition of the induced bias and the analysis of the bias propagated into the model.
    
    To make this analysis possible, we focus on a single large base dataset for FER and use it as a base for creating artificially biased datasets. More specifically, we employ the Affectnet dataset~\cite{Mollahosseini2019}, one of the largest ITW FER datasets, since the large sample size allows us to create biased subsets maintaining a great number of examples. We will create various subsamples of this dataset inducing both representational and stereotypical biases at different strengths. Afterwards, we will train several ResNet50 models~\cite{He2015} and measure how these biases impact the recognition rates of the different emotions for each gender group, making possible to analyze the impact of both representational and stereotypical bias.

    This study aims to explore the propagation of bias in FER systems, from datasets to models, highlighting the risks associated with unchecked data and the need for unbiased data to ensure the fairness and reliability of machine learning models in real-world scenarios. The resulting insights into the propagation of demographic bias in datasets can help shape new bias detection and mitigation strategies, allowing fairer AI systems.

    The rest of the work is structured as follows. First, Section~\ref{sec:related} presents some related work on demographic bias, dataset bias, and FER, as a necessary context for the rest of the work. Next, Section~\ref{sec:method} details the methodology followed, including the implementation details and the overview of the proposed experiments. Section~\ref{sec:results} follows with the results of the experiments, which are then discussed in Section~\ref{sec:discussion}. Finally, Section~\ref{sec:conclusion} provides the conclusions of the work and suggests some lines of future work.

\section{Related work}\label{sec:related}

    In this Section, we review some previous works that serve as a basis to our research. First, Section~\ref{ssec:bias} defines different types of bias and their importance. Then, Section~\ref{ssec:bias} defines the demographic bias metrics used in this work. Finally, Section~\ref{ssec:fer} focuses on FER, our subject problem. 

\subsection{Demographic bias}\label{ssec:bias}

    The recent growth of AI-based systems has raised concerns about their potential for discrimination~\cite{Pessach2020,Landers2022}. To address these concerns, the field of algorithmic justice focuses on conceptualizing and mathematically modeling fairness with the goal of developing practical standards for fair systems. A key approach to fairness is through a negative definition: identifying systems as fair when they lack systematic bias. Bias, in this context, refers to any unjustified and unwanted differences in treatment based on protected demographic attributes~\cite{Suresh2021}. This perception of fairness reflects the legal understanding of bias~\cite{Ntoutsi2020}, which refers to prejudice against individuals or groups based on certain protected demographic attributes. Although the specific legal framework varies across countries, any prejudice based on inherent and immutable attributes, such as race, gender, sex, age, religion, or socioeconomic class, is typically considered protected. Additionally, new regulations and legal frameworks, such as the European AI Act~\cite{Spokesperson2023}, underscore the importance of research on AI bias.

    The potential harm done by AI systems due to inaccurate model predictions can originate from multiple points in the ML pipeline~\cite{Suresh2021}. Nevertheless, the increasing size of the datasets, along with the separation of data gathering and model development teams, make datasets a primary entry points for bias~\cite{Ferrara2024}.

    Bias in image datasets~\cite{Fabbrizzi2022} can manifest in multiple ways, depending on whether it affects the sampled population (selection bias), the assigned target labels (label bias), or the image elements themselves (framing bias). Selection bias is of particular interest because it can affect any dataset that involves human subjects, not just image datasets. This makes research on this variant of bias applicable to a wider array of situations. Previous works~\cite{Dominguez-Catena2024} have focused on this type of bias, establishing metrics and methodologies to measure its variants. This provides a strong foundation for our research on demographic bias propagation. 

    Despite previous work on model and dataset bias, the specific pathways by which dataset bias propagates into the model remain unclear. While most bias mitigation efforts assume that this propagation occurs~\cite{Hort2023}, the limited understanding of how these biases influence the final models motivates the research in this paper.

\subsection{Demographic dataset bias}\label{ssec:biasmetrics}

    Two key variants of demographic dataset bias have been identified~\cite{Dominguez-Catena2024}, representational and stereotypical bias. 
    
    \vspace{0.3em}\noindent\textbf{Representational bias} refers to the skewed presence or absence of demographic groups in a dataset, such as the common overrepresentation of white people in both FER~\cite{Dominguez-Catena2024} and general image datasets~\cite{Dulhanty2019}.

    In this work, we leverage the \textbf{Effective Number of Species (ENS)}, as proposed in ~\cite{Dominguez-Catena2024}, to quantify representational bias. ENS, originally developed in the field of ecology~\cite{Jost2006}, measures the diversity of species within an ecosystem. In the context of bias, the ENS can be roughly interpreted as the number of demographic groups represented. ENS is additionally corrected to consider whether the representation is homogeneous, reducing the final value according to the severity of the imbalances. For example, a dataset with $4$ equally represented groups and an additional underrepresented group will have an ENS between $4$ and $5$. This indicates that it is more representative than a dataset with only $4$ equally represented groups, but less than one with $5$ equally represented ones. Given a dataset $X$, the ENS is defined as follows:
    
    \begin{equation}
        \text{ENS}(X) = \exp\left({-\sum_{g\in G}{p_g \ln p_g}}\right)\;,
    \end{equation}
    where $G$ is the set of possible demographic groups and $p_g$ is the proportion of the dataset of the group $g\in G$.

    This metric has a range of $[1, |G|]$, reaching its maximum value of$|G|$ only when all demographic groups are equally represented.

    In our context, since the number of demographic groups is fixed at $2$ (with gender being one of $G=\{\text{male},\text{female}\}$), we can adopt an alternative notation aligned with the concept of bias, bounded between $0$ (for unbiased datasets) and $1$ (for biased ones):

    \begin{equation}
        \overline{\text{ENS}}(X) = 2 - \text{ENS}(X).
    \end{equation}

    \vspace{0.3em}\noindent\textbf{Stereotypical bias} focuses on the correlation between specific target classes and demographic groups. For example, a common manifestation in FER datasets is the overrepresentation of women in the ``happy'' class~\cite{Dominguez-Catena2024}. In this work, as suggested in ~\cite{Dominguez-Catena2024}, we measure stereotypical bias using \textbf{Cramer's V}~\cite{Cramer1991}. This metric is derived from the Pearson's chi-squared statistic of association (${\chi^2}(X)$), which is the basis for the \textit{Pearson's chi-squared test}. The metric is defined as follows:

    \begin{equation}
        \text{V}(X) = \sqrt{ \frac{\chi^2(X)/n}{\min(|G|-1,|Y|-1)}}\;,
    \end{equation}
    where $n$ is the size of the dataset, $|G|$ the number of possible demographic groups, $|Y|$ the number of target classes and ${\chi^2}(X)$ is defined as
    \begin{equation}
        \chi^2(X)=\sum_{g \in G}\sum_{y \in Y}\frac{(n_{g\land y}-\frac{n_g n_y}{n})^2}{\frac{n_g n_y}{n}}\;,
    \end{equation}
    being $n_g$, $n_y$ and $n_{g\land y}$ the number of samples in each group, target class, or combination of group and target class, respectively.

    The value of $V$ is bounded in $[0, 1]$, where $0$ indicates that there is no stereotypical bias present in the dataset, while $1$ indicates extreme bias.

\subsection{Facial Expression Recognition}\label{ssec:fer}

    Affective Computing is an interdisciplinary field focused on integrating human affect with computational systems. FER is one of the most commonly used modalities within affective computing~\cite{Assuncao2022}, which analyzes visual cues in facial expressions to identify underlying emotions. Using facial images as input offers two key advantages: minimal and easily deployed hardware requirements, and the ability to leverage easily accesible ITW data, such as stock photos, for training. These technologies have various applications, ranging from assistive robotics~\cite{Nimmagadda2022} and healthcare~\cite{Werner2022} to more general human-computer interaction tasks~\cite{Bakariya2023}, including driver drowsiness detection~\cite{Mou2023}. A particularity of FER systems is how they interact with users. Since they are often used to identify the expressions of unsuspecting and non-technical users, they can lead to situations where a user may not be able to recognize a biased system, even if the predictions were clearly wrong for them.
    
    The most extended approach for FER is based on image classification. In this approach, a single face image is classified into a finite set of emotions, usually the six basic emotions proposed by Ekman~\cite{Ekman1971} (\textit{angry}, \textit{disgust}, \textit{fear}, \textit{sad}, \textit{surprise}, and \textit{happy}), often including an additional \textit{neutral} emotion. For specific applications, other sets of emotions and expressions can be used, such as pain~\cite{Werner2022}. However, Ekman-based classification remains the most common in FER datasets~\cite{Goodfellow2013,Zhang2017,Mollahosseini2019} due to its broad understanding and ease of manual labeling.
    
    Since the advent of deep learning, the use of deep convolutional neural networks~\cite{LeCun2015} has become the most common solution for the FER task. One of the most widely used architectures for this task is the residual neural network (ResNet)~\cite{He2015}, which is often used as a baseline. These models tend to generalize well to unseen data and situations. However, training them is expensive and requires large amounts of data. Consequently, most recent FER datasets rely on data obtained from Internet searches~\cite{Nidhi2023}, taking advantage of its diversity and abundance.

    Unfortunately, these \textbf{datasets} are prone to bias issues. Of the two types of bias mentioned in Section~\ref{ssec:bias}, current ITW datasets for FER are more susceptible to stereotypical bias compared to datasets gathered under controlled conditions~\cite{Dominguez-Catena2024}. Controlled datasets typically avoid stereotypical bias by capturing samples of all emotions for each subject. Various demographic axes can be affected by these biases, particularly those that are visually identifiable, such as age, gender, and race. Nevertheless, it is important to remember that facial images can reveal demographics beyond the most obvious ones. For example, in \cite{Bjornsdottir2017} it was shown that the social class can be inferred, potentially contributing to bias dynamics within FER datasets.

    Several recent studies have identified biases related to gender, race, and age within FER \textbf{models}~\cite{Kim2021,Ahmad2022,Xu2020,Domnich2021,Jannat2021,Deuschel2021,Poyiadzi2021}. Kim et al.~\cite{Kim2021} specifically investigate age bias within commercial models, while Ahmad et al.~\cite{Ahmad2022} extend their analysis to include age, gender, and race biases, using a custom database of politician videos. Research models have also been scrutinized for biases. Xu et al.~\cite{Xu2020} investigated age, gender, and race bias in models trained on Internet-sourced data evaluating them using a dataset with known demographic characteristics. Domnich et al.~\cite{Jannat2021} and Jannat et al.~\cite{Domnich2021} specifically investigate gender bias, while Deuschel et al.~\cite{Deuschel2021} focus on biases in predicting facial action units with respect to gender and skin color using popular datasets. Poyiadzi et al.~\cite{Poyiadzi2021} tackle age bias in a dataset gathered from Internet searches. In response to the biases uncovered, researchers have also explored mitigation strategies, including Xu et al.~\cite{Xu2020}, Jannat et al.~\cite{Jannat2021}, and Poyiadzi et al.~\cite{Poyiadzi2021}. Addressing the shortcomings and biases identified in previous studies, Hernandez et al.~\cite{Hernandez2021} propose a set of guidelines aimed at evaluating and mitigating potential risks associated with the application of FER-based technologies.

\section{Methodology}\label{sec:method}

\begin{figure*}
        \centering
        \includegraphics[width=\textwidth]{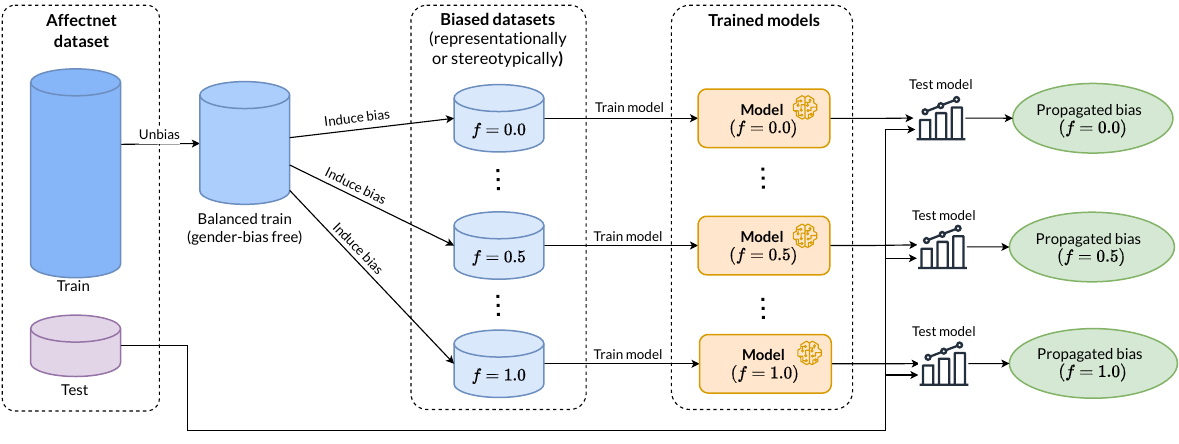}
        \caption{Summary of the experimental methodology.}
        \label{fig:summary}
    \end{figure*}

    This Section presents the methodology employed to study how gender bias propagates from FER datasets to the trained models. To do so, we base our analysis on a single large dataset (summarized in Section \ref{ssec:dataset}), which we use as a base to generate a balanced dataset free from gender bias (both representational and stereotypical). From this bias-free base, we then generate several datasets with induced bias (representational or stereotypical depending on the experiment). Details on the bias induction mechanism are given in Section \ref{ssec:inducedbias}. All generated datasets are used to train several models, which are then evaluated on a test set to analyze potential bias propagation. Details on the training and evaluation processes are given in Section \ref{ssec:setup}. Finally, Section \ref{ssec:experiments} presents the overall details of the experiments conducted. A graphical summary of the complete methodology is presented in Figure~\ref{fig:summary}. 
    

\subsection{Dataset}\label{ssec:dataset}

    From the multiple datasets available for FER~\cite{Nidhi2023}, we chose Affectnet \cite{Mollahosseini2019} for this paper. Our decision is motivated by several reasons. Affectnet is a well-established FER dataset with a relatively large number of face images ($420,299$). This large dataset enables the induction of biases by controlled subsampling of the training partition, while still allowing for accurate models. Additionally, this dataset reflects current trends in ITW datasets, being composed by a large quantity of small, variable-quality images. 
    
    We focus on the main subsection of the dataset labeled according to the basic emotions proposed by Ekman~\cite{Ekman1971}, which comprises a total of $286,341$ images. The rest of the dataset uses a less common valence and arousal codification, with two scalar values. These samples are not used, as it would require an alternative approach to defining and inducing stereotypical bias. We follow the provided division of train and test partitions, which results in $282,853$ train images and $3,488$ test images. The face images in the dataset are in color (RGB) and have a resolution of $425$ by $425$ pixels.

    While the dataset was collected through Internet searches with queries targeting specific emotions and aiming for demographic variety through gender, age, and ethnicity labels, the original data lacks balancing for any demographic property and does not provide demographic labels for the subjects themselves. This absence of demographic information hinders a proper assessment of demographic bias. However, we can address this challenge by following the methodology proposed in previous works~\cite{Dominguez-Catena2023b,Dominguez-Catena2024}, which takes advantage of an auxiliary model to generate proxies for the missing demographic data. In our case, we use FairFace~\cite{Karkkainen2021} as an auxiliary model to estimate the apparent gender from each face image in Affectnet. To apply the model, we first preprocess the images following the procedure carried out for the FairFace training dataset, i.e., applying the Max-Margin Object-Detection (MMOD) CNN face extractor~\cite{King2015} implemented in DLIB\footnote{\url{http://dlib.net/}}. The target image size is $224 \times 224$, with a margin around the face of $56$ pixels. After this, a single binary gender prediction is made for each sample in Affectnet.
    
    Although the FairFace model also offers predictions for apparent age and race, we opted to focus solely on the gender demographics for this work due to several reasons. Firstly, gender classification is a well-established and reliable task within computer vision, and the chosen auxiliary model scores best on this task. Additionally, limiting our analysis to gender and employing a binary classification simplifies the definition of the induced bias and the subsequent analysis of propagation into the model, leading to clearer and more interpretable results. Finally, with both genders being roughly equally represented in the original Affectnet dataset, we can induce identical biases favoring either gender while maximizing datasets size.

    It is important to acknowledge the limitations inherent in using the predicted apparent gender as the demographic attribute. This binary classification of gender may not always align with self-reported gender identity, particularly for individuals who are gender-non-conforming. Despite this caveat, FairFace achieves high accuracy in gender prediction, even on external datasets, with reported performance between $0.92$ and $0.98$~\cite{Karkkainen2021}. This makes it a relatively robust proxy for the real gender in the context of our bias analysis, where we focus on trends across populations rather than individual samples.

    To analyze the gender distribution within Affectnet, we applied the auxiliary model to its images. Table~\ref{tab:cont_affectnet} summarizes this demographic analysis, presenting a contingency table of the predicted apparent gender across the target classes. An additional column shows the proportion of the female group in each of class, highlighting any gender bias in the original dataset.

    \begin{table}[ht]
        \centering
        \begin{tabular}{lrrrr}
        \toprule
          &  \multicolumn{2}{@{}l@{}}{\textbf{Gender}} \\ \cmidrule(r){2-3}
         & Female & Male & F. prop. & Class prop. \\
        \textbf{Class} &  &  & & \\
        \midrule
        angry    &  $6,962$ & $17,803$ & $28.11\%$ & $8.76\%$ \\
        disgust  &  $1,733$ &  $2,054$ & $45.76\%$ & $1.34\%$ \\
        fear     &  $3,117$ &  $3,222$ & $49.17\%$ & $2.24\%$ \\
        happy    & $79,797$ & $54,241$ & $59.53\%$ & $47.39\%$ \\
        neutral  & $33,708$ & $40,858$ & $45.21\%$ & $26.36\%$ \\
        sad      & $11,175$ & $14,156$ & $44.12\%$ & $8.96\%$ \\
        surprise &  $6,439$ &  $7,588$ & $45.90\%$ & $4.96\%$ \\
        \midrule
        Total & $142,931$ & $139,922$ & $50.53\%$ & \\
        \bottomrule
        \end{tabular}
        
        \caption{Original contingency table for the train partition of Affectnet.}
        \label{tab:cont_affectnet}
    \end{table}

    Table~\ref{tab:cont_affectnet} shows that the overall dataset is balanced in terms of gender, with a measured $\overline{\text{ENS}}$ of \num{1.13e-4}. However, balancing is not maintained across labels, showing significant imbalances in labels such as ``happy'' (60\% female) and ``angry'' (less than 30\% female). This is an example of the stereotypical bias present in ITW datasets~\cite{Dominguez-Catena2024}. As a consequence, measuring stereotypical bias with Cramer's V leads to $\text{V}=0.19$. Despite these biases, the dataset is large enough to allow us to obtain a balanced subset on each label in order to induce biases, as described in the next section.

\subsection{Bias Induction and Measurement}\label{ssec:inducedbias}

    To fairly compare representational and stereotypical bias, we artificially induce gender bias in the train partition of the dataset while maintaining the test partition intact. This involves adopting a systematic subsampling approach from the original dataset to strategically manipulate the gender proportions (overall or within specific classes). By controlling the gender proportions within classes, we aim to simulate scenarios where certain demographic groups are overrepresented or underrepresented either in the whole dataset (representational bias) or in specific classes (stereotypical bias). Models trained under either scenario can then reflect the effects of gender bias on model performance.

    
    The specific bias induction technique consists of two phases. First, a balanced baseline is generated with a $50\%$ ratio of each gender group for all classes; then, specific biased datasets are generated for each bias type (either representational or stereotypical). In the following, we provide a detailed description of these phases and how we apply them to generate each type of bias.
    
    \begin{enumerate}
        \item \textbf{Generation of a balanced baseline.} 
        To avoid any biases in the baseline dataset, we subsample the largest possible gender balanced subsample for each class. To do so, we determine the amount of samples of the least represented gender group in each class ($n_c$) and subsample the overrepresented gender group in each class to the same amount.  This process slightly modifies the original class distribution of Affectnet, but aiming to maintain the class distribution would imply losing a significant amount of samples. The summary of the resulting dataset as a contingency table is provided in Table~\ref{tab:cont_balanced}.

    \begin{table}[ht]
        \centering
        \begin{tabular}{lrrrr}
        \toprule
          &  \multicolumn{2}{@{}l@{}}{\textbf{Gender}} \\ \cmidrule(r){2-3}
         & Female & Male & F. prop. & Class prop. \\
        \textbf{Class} &  &  & &  \\
        \midrule
        angry    & $6,962$ & $6,962$ & $50\%$ & $5.93\%$ \\
        disgust  & $1,733$ & $1,733$ & $50\%$ & $1.48\%$ \\
        fear     & $3,117$ & $3,117$ & $50\%$ & $2.66\%$ \\
        happy    & $54,241$ & $54,241$ & $50\%$ & $46.21\%$ \\
        neutral  & $33,708$ & $33,708$ & $50\%$ & $28.72\%$ \\
        sad      & $11,175$ & $11,175$ & $50\%$ & $9.52\%$ \\
        surprise & $6,439$ & $6,439$ & $50\%$ & $5.49\%$ \\
        \midrule
        Total    & $117,375$ & $117,375$ & $50\%$ & \\
        \bottomrule
        \end{tabular}
        
        \caption{Contingency table for the balanced train partition used as a source for the biased datasets.}
        \label{tab:cont_balanced}
    \end{table}
    
        \item \textbf{Bias induction.} For each class $c$, we define a bias factor $f_c\in[0,1]$ to represent the desired proportion of female samples in class $c$. We then subsample the female group in that class to a total of $n_c \times f_c$ and the male group to a total of $n_c \times (1 - f_c)$. This approach ensures that all generated datasets have the same size (half of the baseline dataset), preventing potential biases that could arise due to differences in dataset size.
    \end{enumerate}

    Depending on the $f_c$ values defined for each class, we can simulate both representational and stereotypical bias scenarios:

    \begin{itemize}
        \item \textbf{Representational bias} can be induced by setting the same value $f_c=f$ for every class $c$. This results in a global proportion of $f$ female subjects in the dataset, ensuring equal representation across all classes. Consequently, we reduce the parameterization of representational bias to a single bias factor~$f$.
        \item \textbf{Stereotypical bias} can be induced by manipulating the $f_c$ of a single class $c$, while setting the bias factors for all other classes $c'$ to $f_{c'}=0.5$. While this approach introduces a slight representational bias (since the female-to-male ratio varies in the targeted class only), it ensures that other classes remain unmodified. This is a deliberate choice to isolate the impact of stereotypical bias on a single class.
        
        Although stereotypical bias can affect multiple classes simultaneously, we focus on the prototypical single-class variant in this work, leaving more complex scenarios for future research. We parameterize stereotypical bias by both the target class $c$ and the applied bias factor~$f_{c}$. Since we only modify one bias factor $f_{c}$, we will refer to it as $f$ and assume the default $f_{c'}=0.5$ for all other classes.
    \end{itemize}
    
\subsection{Experimental Setup}\label{ssec:setup}

    For each biased dataset generated, we train a ResNet50 network~\cite{He2015} initialized with ImageNet-1K pretrained weights provided by the PyTorch library. This architecture is a standard choice for image classification, and the readily available PyTorch weights establish it as a common baseline for computer vision projects. Training proceeds for $20$ epochs, using a 1cycle policy~\cite{Smith2018} with a maximum learning rate of $1e^{-4}$.

    After training, the models are evaluated on the entire Affectnet test set. For this evaluation, we will go beyond using the overall model acuracy. Instead, we calculate the recall for each target class, independently for each gender group. This approach allows us to identify bias by analyzing the discrepancy in the recall between male and female groups, while remaining robust to the inherent variations in recall across the different target classes. 
    
    To mitigate the influence of randomness inherent in both subsampling and training, we repeat the biased dataset generation, training, and evaluation processes three times for each configuration. Subsequently, we average the recall values obtained across these three runs.

    The experiments are developed using PyTorch 2.0.1. The hardware setup consists of a PC equipped with a GeForce RTX 2060 Super GPU, 20~GB of RAM and an Intel\textregistered\ Xeon\textregistered\ i5-8500 CPU running the Ubuntu Linux 22.04 operating system.
    
\subsection{Overview of experiments}\label{ssec:experiments}

    To better understand how both representational and stereotypical bias propagate to the model, we propose two experiments, one for each type of bias.

    \begin{itemize}
        \item \textit{Experiment 1. Representational bias.} In this experiment, we investigate how manipulating the global representation balance between genders influences the predictions of the final model. Common intuition suggests that underrepresenting a particular gender should worsen the model's ability to recognize its expressions effectively, resulting in lower class recalls for the underrepresented group compared to the overrepresented one.

        For this, we train models on representationally biased datasets with bias factors ($f$) ranging from $0$ to $1$, in increments of $0.1$. This results in $11$ datasets, including a reference balanced dataset at $f=0.5$.
        
        \item \textit{Experiment 2. Stereotypical bias.} In this experiment, we investigate the impact of manipulating the gender balance within a single class, while maintaining balance in all other classes. We expect the recognition of the manipulated class to degrade for the underrepresented group, leading to a lower recall. The impact on the recalls of other classes, that is, how localized is the effect of stereotypical bias, remains unclear. However, prior research~\cite{Dominguez-Catena2023} suggests minimal cross-class interference.

        For this experiment, we generate seven sets of stereotypically biased datasets, one for each target class. Each set contains $11$ datasets with bias factors ranging from $0$ to $1$ in $0.1$ increments. We train a model on each individual dataset, resulting in a total of $77$ datasets for the experiment.
    \end{itemize}

\section{Results}\label{sec:results}

    In this section, we begin by presenting the dataset bias metrics of the generated biased datasets. These metrics, presented in Section~\ref{ssec:res_eval}, validate the biasing process. Subsequently, Sections~\ref{ssec:res_rep} and~\ref{ssec:res_stereo} detail the results of the two experiments, addressing representationally and stereotypically biased datasets and their corresponding models, respectively.
    

    \subsection{Evaluation of induced bias}\label{ssec:res_eval}

    After applying the subsampling process described in Section~\ref{ssec:inducedbias}, we generated a total of 88 training datasets with varying bias distributions for both representational and stereotypical biases. Three representative datasets from each type are presented in Table~\ref{tab:cont_rep} (representational bias) and Table~\ref{tab:cont_stereo} (stereotypical bias, for brevity, just in the angry class) for reference. These examples showcase datasets for the balanced reference ($f=0.5$) and the two extremes of bias induction ($f=0$ and $f=1$). Note that in representational bias, the two extremes involve keeping the maximum number of samples from one group while eliminating the other (e.g., all male or all female). In stereotypical bias, the process is similar but limited to the single targeted class (e.g., angry faces). However, in the latter case, some representational bias is also unintentionally introduced, proportional to the number of samples in the biased target class (as seen in the "Total" row of the table).


    \begin{table}[ht]
        \centering
        \resizebox{\columnwidth}{!}{
        \begin{tabular}{@{}lrrrrrr@{}}
            \toprule
            \textbf{Bias factor} & \multicolumn{2}{l@{}}{$f = 0.0$} & \multicolumn{2}{l@{}}{$f = 0.5$} & \multicolumn{2}{l@{}}{$f = 1.0$} \\
            \cmidrule(rl){2-3} \cmidrule(rl){4-5} \cmidrule(rl){6-7}
            \multicolumn{1}{@{}r@{}}{\textbf{Gender}} &    Female & Male & Female &   Male &  Female & Male \\
            \textbf{Class}    &         &        &        &        &         &      \\
            \midrule
            angry & 0 & 6,962 & 3,481 & 3,481 & 6,962 & 0 \\
            disgust & 0 & 1,733 & 866 & 866 & 1,733 & 0 \\
            fear & 0 & 3,117 & 1,558 & 1,558 & 3,117 & 0 \\
            happy & 0 & 54,241 & 27,120 & 27,120 & 54,241 & 0 \\
            neutral & 0 & 33,708 & 16,854 & 16,854 & 33,708 & 0 \\
            sad & 0 & 11,175 & 5,587 & 5,587 & 11,175 & 0 \\
            surprise & 0 & 6,439 & 3,219 & 3,219 & 6,439 & 0 \\
            \midrule
            Total & 0 & 117,375 & 58,685 & 58,685 & 117,375 & 0 \\
            \bottomrule
        \end{tabular}}
        \caption{Contingency tables for three datasets with varying degrees of representational bias.}
        \label{tab:cont_rep}
    \end{table}

    \begin{table}[ht]
        \centering
        \resizebox{\columnwidth}{!}{
        \begin{tabular}{lrrrrrr}
            \toprule
            \textbf{Bias factor} & \multicolumn{2}{l@{}}{$f = 0.0$} & \multicolumn{2}{l@{}}{$f = 0.5$} & \multicolumn{2}{l@{}}{$f = 1.0$} \\
            \cmidrule(rl){2-3} \cmidrule(rl){4-5} \cmidrule(rl){6-7}
            \multicolumn{1}{@{}r@{}}{\textbf{Gender}} & Female &   Male & Female &   Male & Female &   Male \\
            \textbf{Class}    &        &        &        &        &        &        \\
            \midrule
            angry & 0 & 6,962 & 3,481 & 3,481 & 6,962 & 0 \\
            disgust & 866 & 866 & 866 & 866 & 866 & 866 \\
            fear & 1,558 & 1,558 & 1,558 & 1,558 & 1,558 & 1,558 \\
            happy & 27,120 & 27,120 & 27,120 & 27,120 & 27,120 & 27,120 \\
            neutral & 16,854 & 16,854 & 16,854 & 16,854 & 16,854 & 16,854 \\
            sad & 5,587 & 5,587 & 5,587 & 5,587 & 5,587 & 5,587 \\
            surprise & 3,219 & 3,219 & 3,219 & 3,219 & 3,219 & 3,219 \\
            \midrule
            Total & 55,204 & 62,166 & 58,685 & 58,685 & 62,166 & 55,204 \\
            \bottomrule
        \end{tabular}}
        \caption{Contingency tables for three datasets with varying degrees of stereotypical bias targeting the \textit{angry} class.}
        \label{tab:cont_stereo}
    \end{table}

\begin{figure*}[ht]
        \centering
        \includegraphics[width=\linewidth]{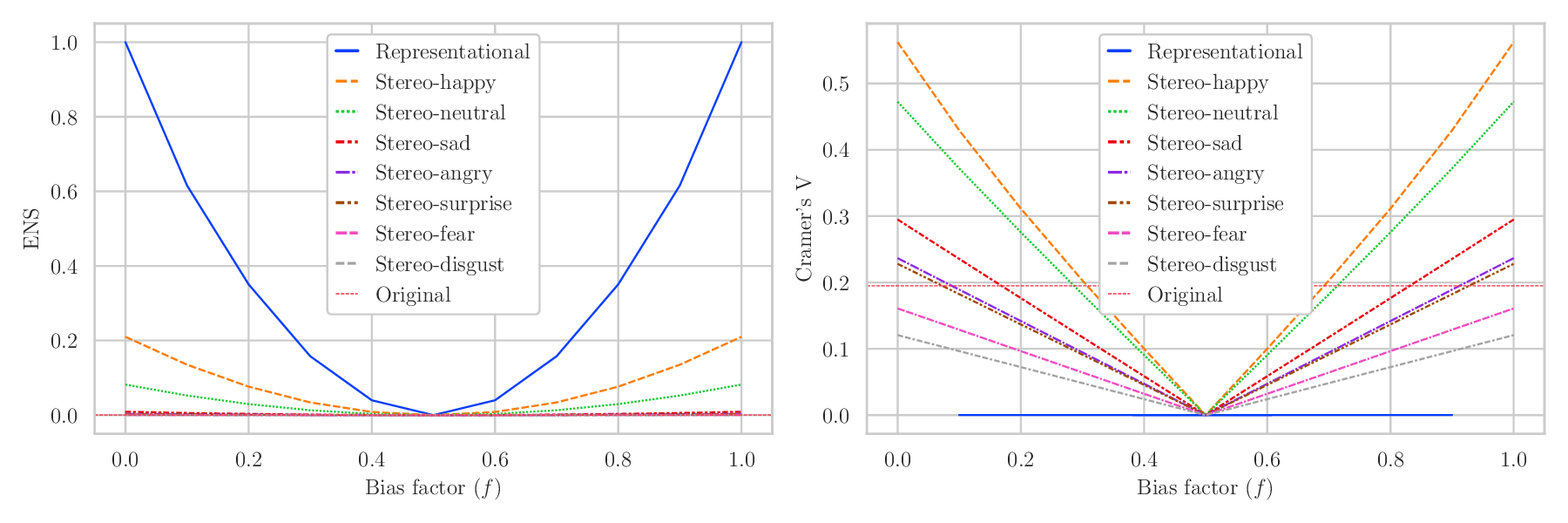}
        \caption{Dataset bias as measured with $\overline{\text{ENS}}$  (representational bias, left) and Cramer's $\text{V}$ (stereotypical bias, right), for datasets with different degrees of induced bias. In both plots, the horizontal red lines represent the bias metrics of the original Affectnet dataset.}
        \label{fig:dataset_bias}
    \end{figure*}

    To quantify the amount of bias generated in each dataset, we employed the two bias metrics presented in Section~\ref{ssec:biasmetrics}. Figure~\ref{fig:dataset_bias} shows the results of applying these metrics to the biased datasets, with $\overline{\text{ENS}}$ measuring representational bias and Cramer's $\text{V}$ measuring stereotypical bias. Importantly, these metrics do not consider which gender class is under or overrepresented, whereas our factor $f$ does, making the relationship symmetric around the reference point at $f=0.5$.

    Regarding \textbf{representational bias} (measured by $\overline{\text{ENS}}$), the values achieved by representationally biased datasets indicate that a significant amount of representational bias is induced, reaching its theoretical maximum ($1$ for a binary demographic attribute). Additionally, as expected, the stereotypically biased datasets exhibit a slight representational bias side-effect. This peaks at $0.2$ when biasing the \textit{happy} class, which has the largest number of samples and thus the greatest impact in generating representational bias.
    
    Focusing on \textbf{stereotypical bias} (measure by Cramer's V), the amount of generated stereotypical bias depends on the relative size of the corresponding class. Accordingly, the largest class (\textit{happy}) achieves the strongest measured induced bias $0.56$), with the theoretical maximum for this metric being $1$. Of course, representationally biased datasets show no stereotypical bias ($\text{V} = 0$).

    Overall, these results demonstrate the validity of the bias induction process, in line with previous definitions of both representational and stereotypical bias.

    \subsection{Experiment 1: Bias propagation in representational bias}\label{ssec:res_rep}

\begin{figure}[htbp]
        \centering
        \subfloat[\textbf{Difference in recalls} (F-M) under representational bias]{\includegraphics[width=\columnwidth]{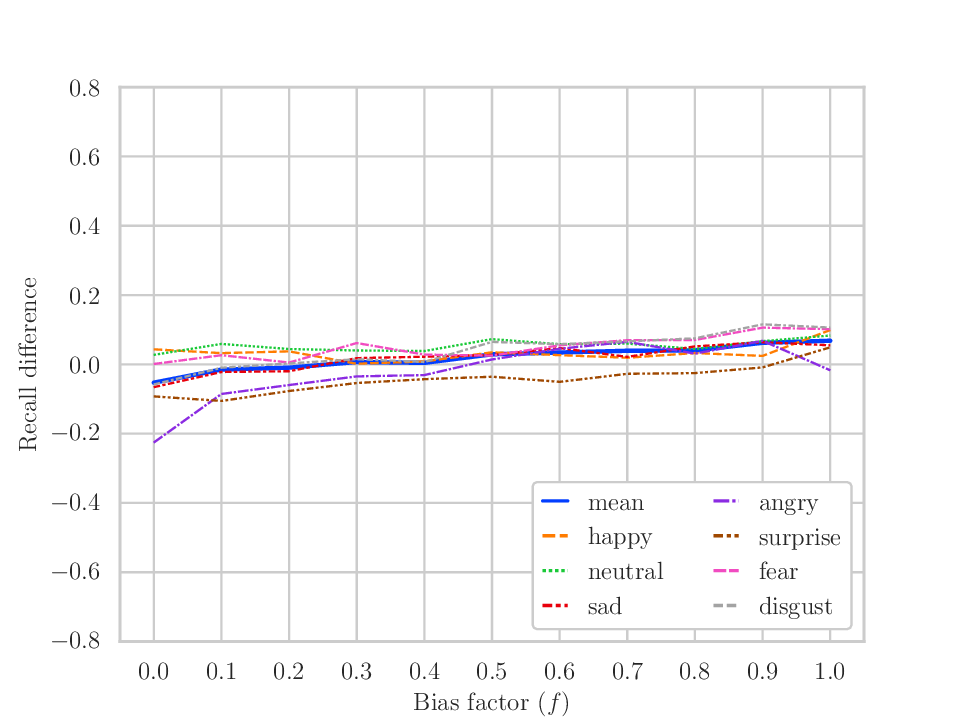}\label{sfig:recalldiffs_representational}}
        \vspace{-.7em}
        \subfloat[\textbf{Recall for the female group} under representational bias]{\includegraphics[width=\columnwidth]{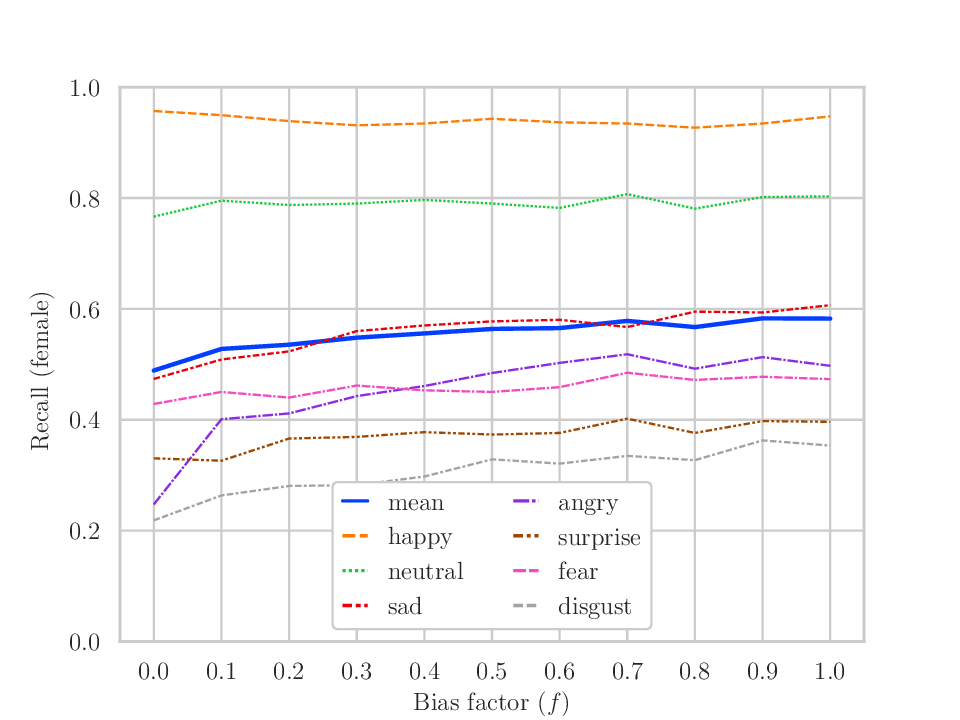}\label{sfig:recallfem_representational}}
        \vspace{-.7em}
        \subfloat[\textbf{Recall for the male group} under representational bias]{\includegraphics[width=\columnwidth]{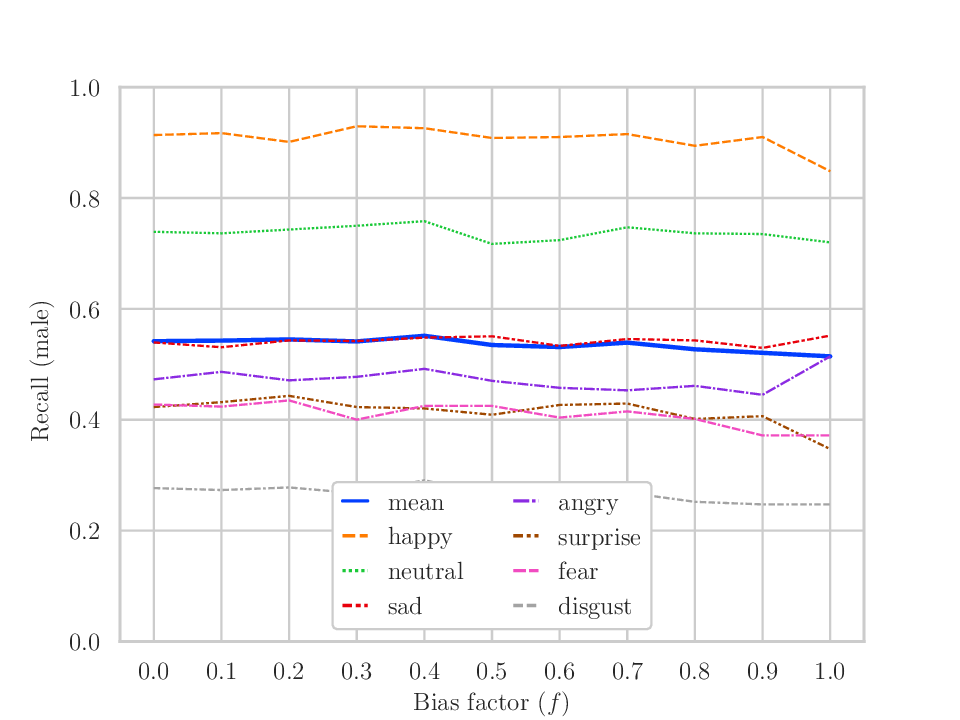}\label{sfig:recallmale_representational}}

        \caption{(a) Recall difference (female recall minus male recall) for the representationally biased datasets. (b) Recall per class for the female group. (c) Recall per class for the male group. For all three, in the horizontal axis, amount of induced bias (bias factor $f$).}
        \label{fig:recalldiffs_representational}
    \end{figure}

    The performance of the models trained on representationally biased datasets is summarized in Figure~\ref{fig:recalldiffs_representational}. Figure~\ref{sfig:recalldiffs_representational} shows the difference between the female and male recalls for varying levels of bias and for each individual target emotion, while Figures~\ref{sfig:recallfem_representational} and~\ref{sfig:recallmale_representational} show the recalls for the female and male groups, respectively.

    Focusing on the recall differences (Figure~\ref{sfig:recalldiffs_representational}), we observe that the differences remain close to $0$ for all target emotions, even in cases of extreme bias ($f=0$ and $f=1$). The strongest difference occurs for the \textit{angry} class when female examples are removed ($f=0$), resulting in a difference of $0.24$ to the disadvantage of the female group. Additionally, the overall effect of representational bias can be observed in the \textit{mean} line (a macro average of the rest of the class-recall differences). For this type of bias, recalls degrade on average from $-0.06$ for $f=0$ to $0.07$ for $f=1$, showing an overall weak propagation.
    
    If we further explore the source of the differences in recalls (Figures~\ref{sfig:recallfem_representational} and~\ref{sfig:recallmale_representational}), it seems that they can be explained by lower recalls when a group is underrepresented ($f<0.5$ for female in Figure~\ref{sfig:recallfem_representational} and $f>0.5$ for male in Figure~\ref{sfig:recallmale_representational}), rather than any potential improvements when they are overrepresented. Interestingly, the most extreme case (\textit{angry} class in $f=0$), where female recall suffers greatly from underrepresentation in the training data, does not replicate when the underrepresented group is the male group ($f=1$), which seems mostly unaffected by underrepresentation.

    \subsection{Experiment 2: Bias propagation in stereotypical bias}\label{ssec:res_stereo}

    In this experiment and due to the number of target emotions, we focus on the \textit{angry} class first. We will then provide a global analysis for the remaining classes (detailed results in Appendix~\ref{app:gender_disagg}).

\begin{figure}[htbp]
        \centering
        \subfloat[\textbf{Difference in recalls} (F-M) under stereotypical bias (angry class)]{\includegraphics[width=\columnwidth]{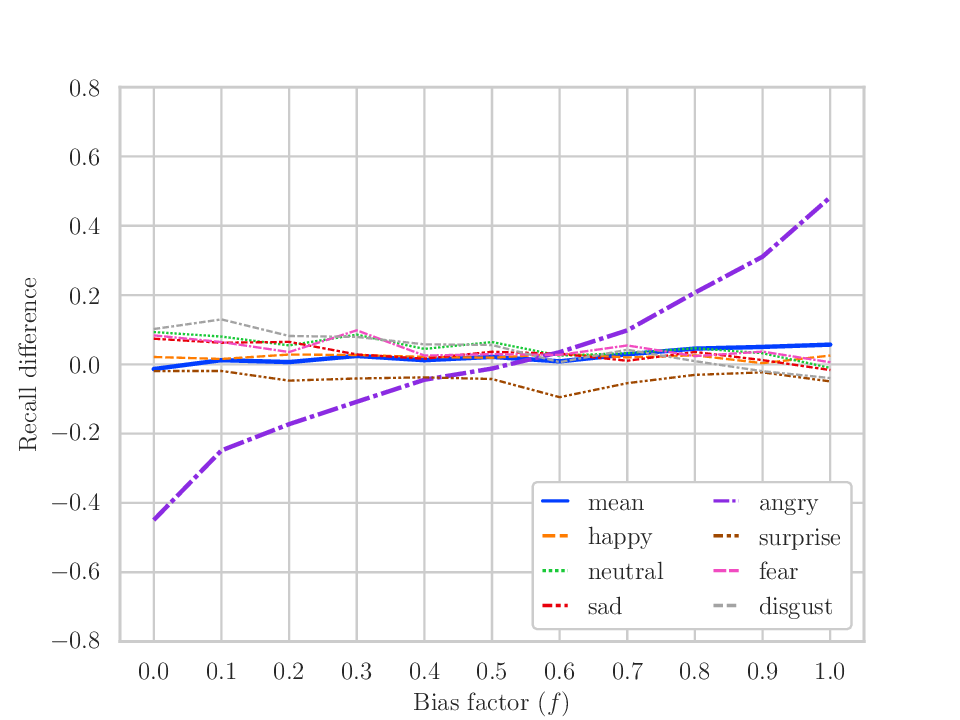}\label{sfig:recalldiffs_angry}}
        \vspace{-.7em}
        \subfloat[\textbf{Recall for the female group} under stereotypical bias (angry class)]{\includegraphics[width=\columnwidth]{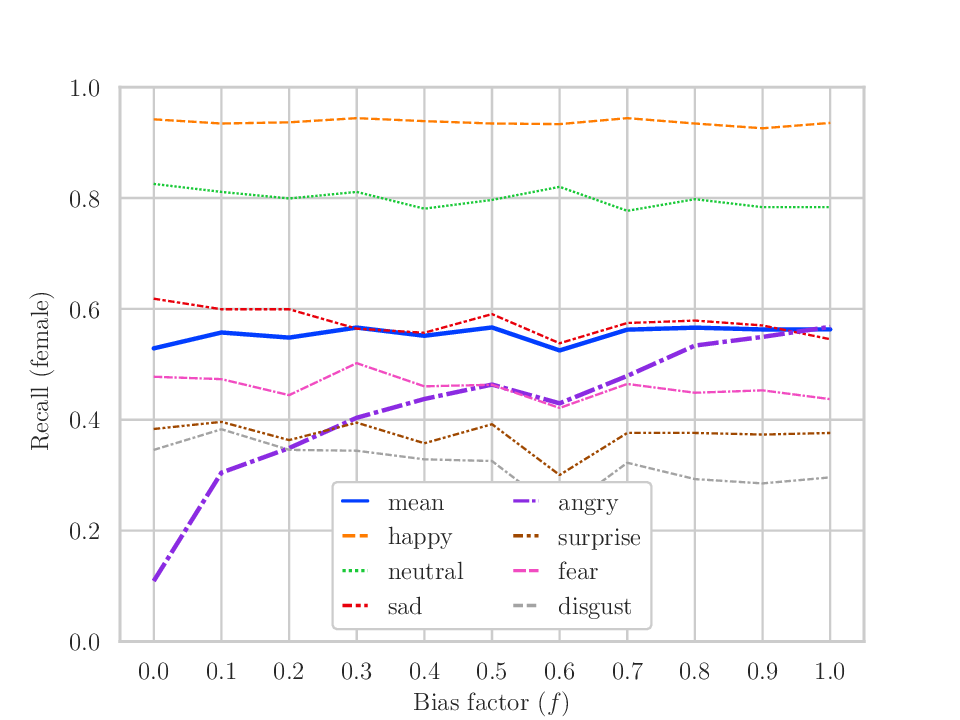}\label{sfig:recallfem_angry}}
        \vspace{-.7em}
        \subfloat[\textbf{Recall for the male group} under stereotypical bias (angry class)]{\includegraphics[width=\columnwidth]{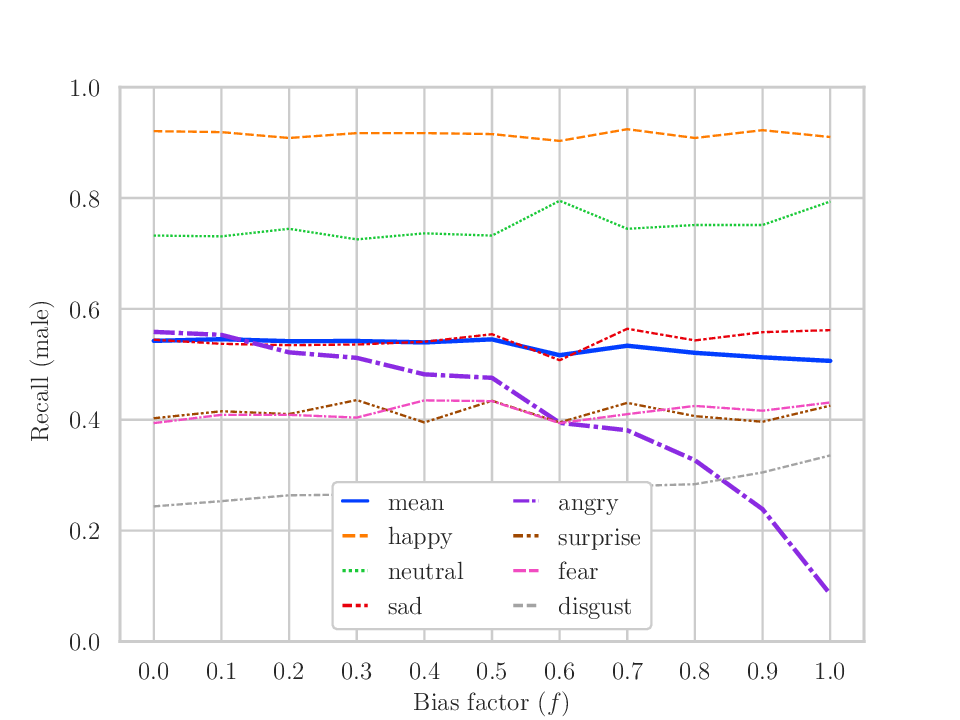}\label{sfig:recallmale_angry}}

        \caption{(a) Recall difference (female recall minus male recall) for each stereotypically biased dataset with biased class \textit{angry}. (b) Recall per class for the female group. (c) Recall per class for the male group. For all three, in the horizontal axis, amount of induced bias (bias factor $f$)..}
        \label{fig:recalldiffs_angry}
    \end{figure}

    Following the previous experiment's structure, Figure~\ref{fig:recalldiffs_angry} shows the results for the stereotypically biased datasets targeting the \textit{angry} class. 
    Recall differences (female to male) across all classes, including the biased \textit{angry} class, are shown in Figure~\ref{sfig:recalldiffs_angry}. Unlike models trained on representationally biased datasets, the propagated bias effect on the angry class is evident. Under extreme bias conditions ($f=0$ and $f=1$), we observe a remarkable difference in female and male recalls, ranging from $-44\%$ to $47\%$. 

    Looking at the dissagregated results (female and male recalls) in Figures~\ref{sfig:recallfem_angry} and~\ref{sfig:recallmale_angry}, we observe several trends. In the overrepresented groups ($f>0.5$ for female and $f<0.5$ for male) there is a weak increase in recall (under $0.1$). Conversely, the impact is more pronounced in the underrepresented groups ($f>0.5$ for female and $f<0.5$ for male), with a total drop of up to $0.4$ in the extreme cases compared to the balanced reference ($f=0.5$). Additionally, the effect of stereotypical bias is stronger for the biased class, yet also noticeable in the rest of the classes. Finally, other emotions show variations of less than $\pm0.1$, generally opposing the variations observed in the \textit{angry} class.

\begin{figure*}[htbp]
        \centering
        \subfloat[Difference in recalls (F-M) under stereotypical bias (\textbf{disgust} class)]{\includegraphics[width=0.48\linewidth]{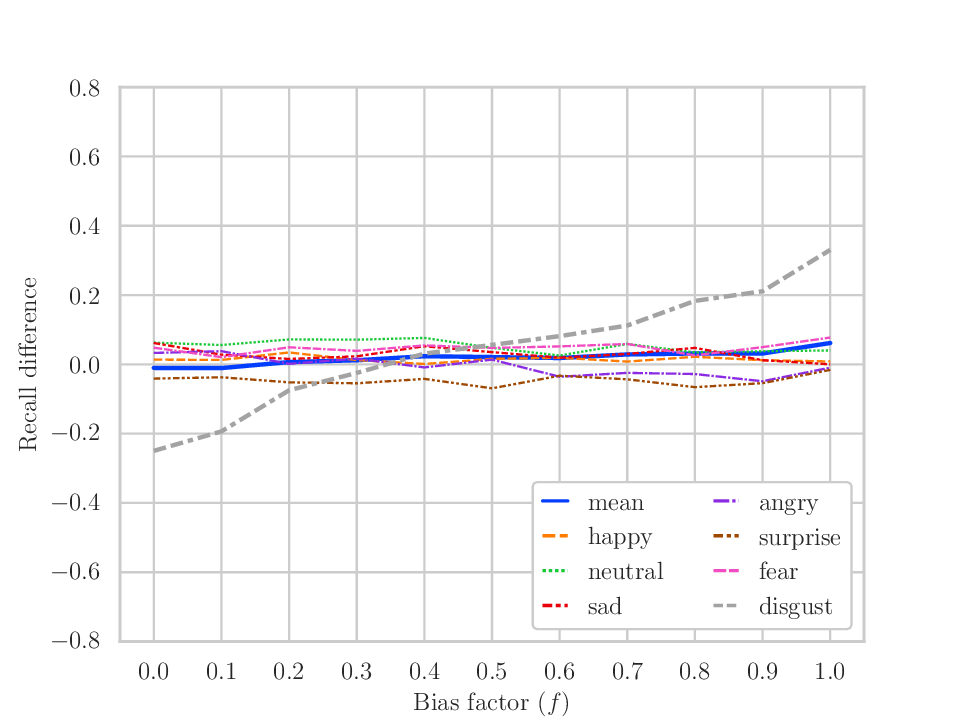}\label{sfig:recalldiffs_disgust}}
        \subfloat[Difference in recalls (F-M) under stereotypical bias (\textbf{fear} class)]{\includegraphics[width=0.48\linewidth]{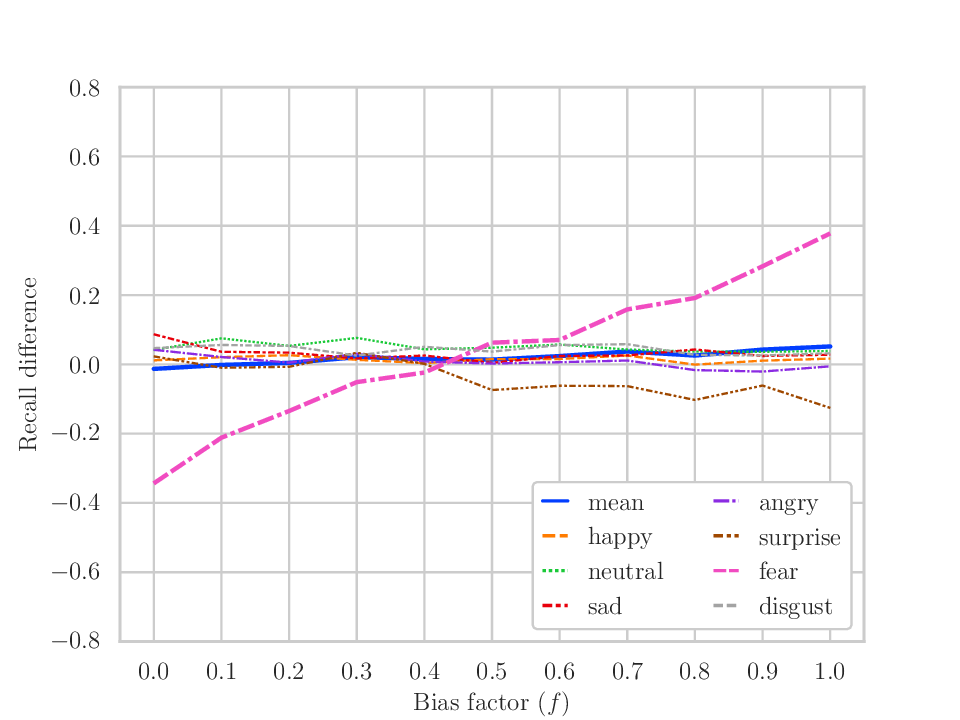}\label{sfig:recalldiffs_fear}}
        \vspace{-.7em}
        \subfloat[Difference in recalls (F-M) under stereotypical bias (\textbf{happy} class)]{\includegraphics[width=0.48\linewidth]{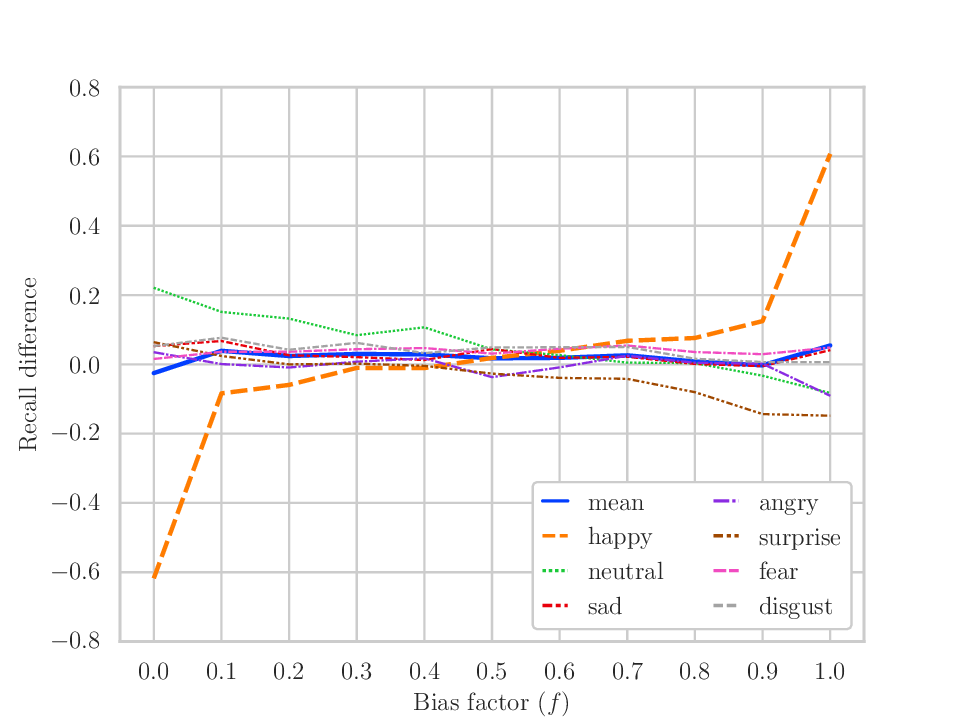}\label{sfig:recalldiffs_happy}}
        \subfloat[Difference in recalls (F-M) under stereotypical bias (\textbf{neutral} class)]{\includegraphics[width=0.48\linewidth]{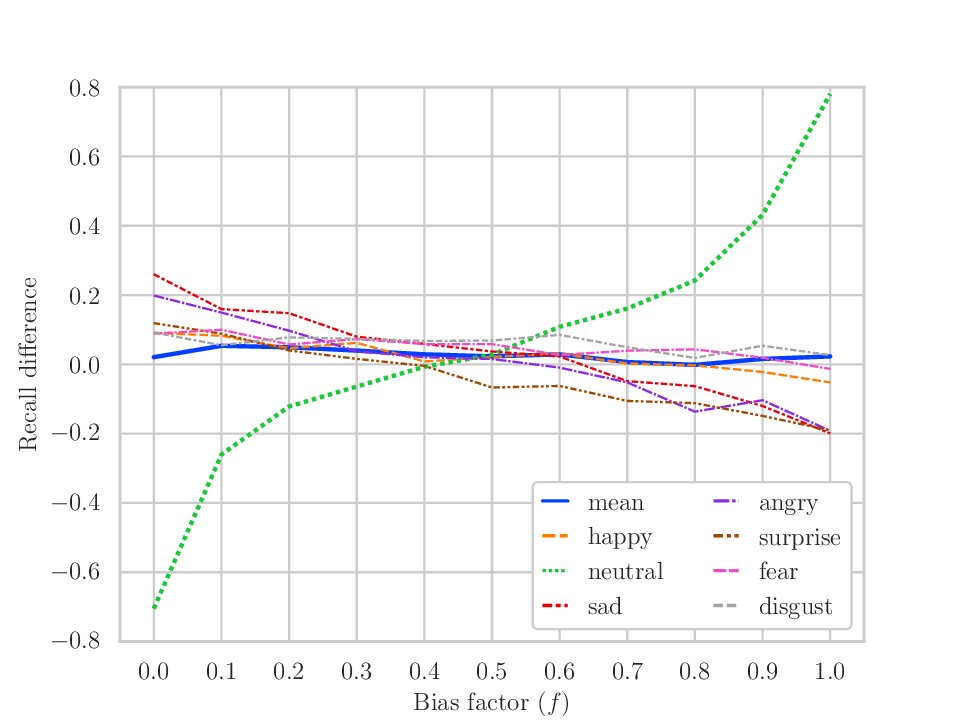}\label{sfig:recalldiffs_neutral}}
        \vspace{-.7em}
        \subfloat[Difference in recalls (F-M) under stereotypical bias (\textbf{sad} class)]{\includegraphics[width=0.48\linewidth]{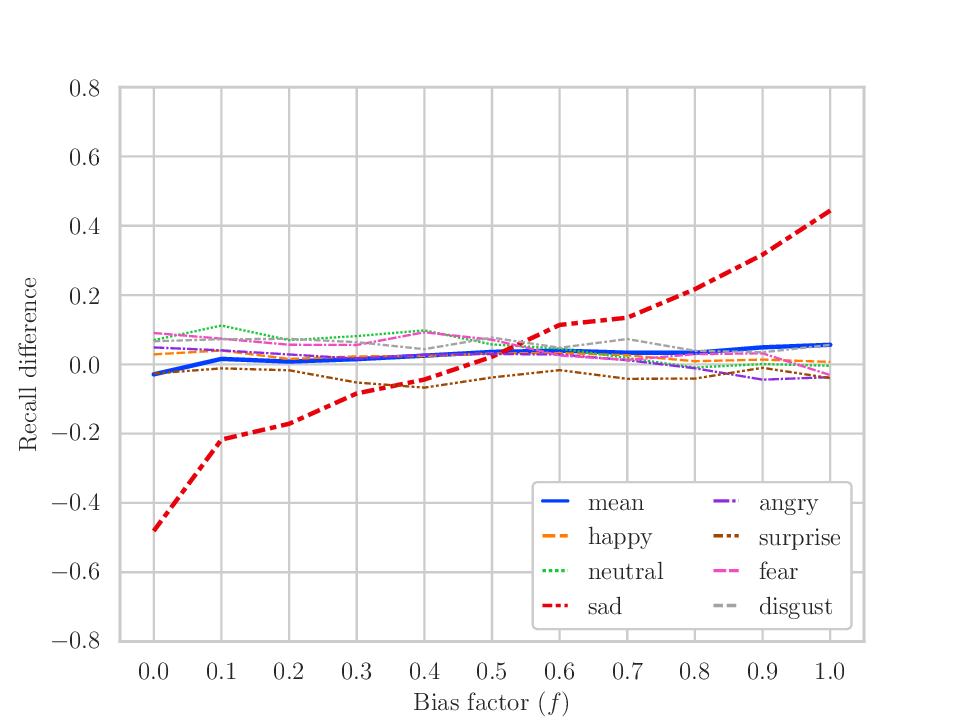}\label{sfig:recalldiffs_sad}}
        \subfloat[Difference in recalls (F-M) under stereotypical bias (\textbf{surprise} class)]{\includegraphics[width=0.48\linewidth]{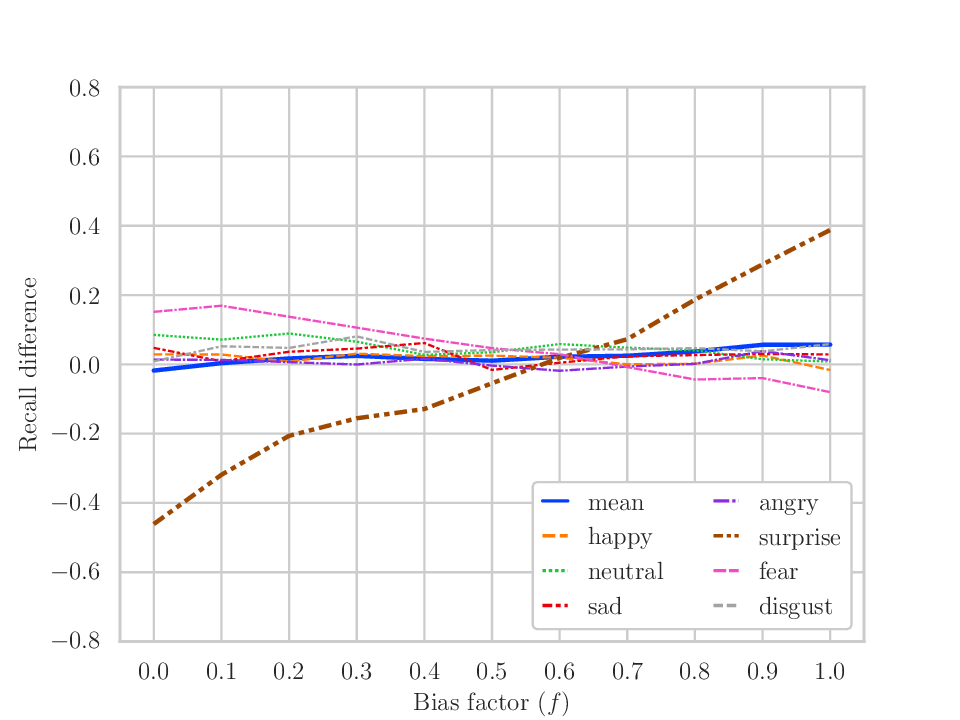}\label{sfig:recalldiffs_surprise}}
        \caption{Class recall differences (female to male) across six sets of stereotypically biased datasets, targeting \textit{disgust}, \textit{fear}, \textit{happy}, \textit{neutral}, \textit{sad} and \textit{surprise} classes. In all subplots, the horizontal axis represents the amount of induced bias (bias factor $f$).}
        \label{fig:recalldiffs}
    \end{figure*}

    For the stereotypical bias induction on the other six emotions, the difference in recalls are shown in Figure~\ref{fig:recalldiffs}. Each graph shows the difference in recall (female to male) for the stereotypically biased datasets in each emotion, with the horizontal axis on each plot corresponding to the strength of the induced bias (bias factor $f$). Recall that the dissagregated results are provided in Appendix~\ref{app:gender_disagg}.

    Similar to the angry class, all cases show a substantial effect from the stereotypically induced bias, with this effect being significantly stronger within the biased class itself. The most extreme example is the \textit{neutral} class, where a difference in recalls of $-0.69$ is observed at $f=0$  and a difference of $0.76$ at $f=1$. Bias propagation is largely linear for non-extreme values of $f$ ($f\in[0.1,0.9]$), but exhibits a tendency to intensify towards the extremes. Two emotions (\textit{happy} and \textit{neutral}) exhibit an abrupt change in recall difference for $f=0$ and $f=1$, suggesting an intensification of the bias effect. This intensification results in a female-male difference of up to $\pm0.6$. The remaining emotions show a more linear response, with recall differences ranging between $\pm0.2$ and $\pm0.5$ in extreme bias cases.
    

    For most stereotypically biased datasets, the propagated bias primarily affects the biased class itself. This holds true for \textit{angry}, \textit{disgust}, \textit{fear}, \textit{sad} and \textit{surprise}. However, \textit{happy} and \textit{neutral}, which exhibit a greater overall impact from stereotypical bias, demonstrate a strong effect on both the biased class and the remaining classes. When biasing these two emotions, other classes register variations of up to $\pm0.2$ in extreme cases. This is comparable to, and even exceeds, the variations observed with representational bias (Figure~\ref{sfig:recalldiffs_representational}). Note that this effect also correlates with the class sizes, with \textit{happy} and \textit{neutral} being the most frequent classes (and thus incurring more representational bias during stereotypical bias induction, as shown in Section~\ref{ssec:res_eval}).
    


\section{Discussion}\label{sec:discussion}

    Overall, the impact of \textbf{representational bias} is unexpectedly weak. Even in extreme cases where one gender group is entirely omitted, the model demonstrates reasonable generalization to the other group. Only one of the less supported classes (\textit{angry}) experiences a significant impact from the imposed bias. Interestingly, generalization appears more successful for one group over the other; omitting male samples has a lesser impact than omitting female ones. This asymmetry could stem from inherent differences in the way males and females express emotions, which has been previously studied~\cite{Atkinson2005}. However, further research is needed to confirm this hypothesis.

    Compared to representational bias, the propagation of \textbf{stereotypical bias} is more pronounced. This is evident in the recall differences, which primarily reflect a decrease in recall for the underrepresented group, with minimal gain for the overrepresented group. The effect varies across classes, influenced by both the size and difficulty of the class. For instance, the \textit{happy} and \textit{neutral} classes exhibit the largest impact on recall differences. These classes are also the most frequent in the dataset, leading to the highest measured stereotypical bias (using Cramer's V). This likely explains their magnified impact. However, it is important to note that these classes also seem easier to recognize correctly. Our intuition is that this ease stems from two distinct sources. The \textit{happy} class is naturally easier to identify due to clear visual features like a smile and visible teeth, which makes it learnable even with fewer examples. Conversely, the \textit{neutral} class might not be as visually distinctive, but its large size provides enough data for the model to achieve high recognition accuracy. Hence, our experiments cannot definitively rule out these issues as a contributing factor to the observed bias propagation.


    Stereotypical bias propagation seems largely linear and localized to the biased class. However, in \textit{happy} and \textit{neutral} emotions, this effect intensifies at the extremes ($f=0$ and $f=1$). Additionally, these same cases exhibit bias leakage to other emotions, evidenced by recall differences showing a weaker, opposing trend to that of the biased class. Our hypothesis, building on our previous explanation, is that completely removing a gender group from these classes leads the model to classify examples of that group more readily into the other classes, thereby improving the recall for those classes. This suggests the model learns a new gender-related pattern: to minimize the classification of the samples from the missing group into the biased emotion. Notably, this pattern is more pronounced in the extremes. For the \textit{happy} class, likely because it is inherently easier to recognize and requires fewer examples for good performance. Conversely, the neutral class exhibits a more linear behavior, probably due to its higher recognition difficulty, requiring more examples to maintain the performance. 

    We further hypothesize that the lower impact of bias propagation in unbiased classes when biasing the rest of the classes (\textit{angry}, \textit{disgust}, \textit{fear}, \textit{sad} and \textit{surprise}) also stems from their inherent difficulty and baseline recall values. Since models typically achieve recall values below $0.5$ for these classes (due to either inherent difficulty and the availability of fewer training examples), even if the model learns the pattern of avoiding the biased emotion for a specific gender group, the effect might be less pronounced. This is likely because the induced bias itself is weaker in these classes due to the few training examples available. Nonetheless, further research is necessary to validate these hypotheses.

    Although our findings generally align with prior research on stereotypical bias propagation~\cite{Dominguez-Catena2023}, the observed impact on the models appears stronger and without any safe zone. That is, small amounts of stereotypical bias seem to affect model predictions. This suggests that dataset size may play a role in the intensity of stereotypical bias propagation. Our significantly larger training dataset ($117,375$ images) compared to the FER+ dataset ($9,475$ images) used in~\cite{Dominguez-Catena2023} supports this hypothesis. However, further experiments are necessary to confirm this relationship.


   The stronger observed effect of stereotypical bias compared to representational bias can likely be attributed to the training process itself. In representational bias, where a demographic group is underrepresented across all classes, the model can often generalize from the overrepresented group to the underrepresented one, assuming similar emotional expressions between the groups. However, stereotypical bias induces a new pattern that can be learned by the model: the observable characteristics of a gender group can be used to improve the accuracy of the biased training data. This also explains the increase in recall for the unbiased classes in the underrepresented group, since taking advantage of the gender appearance to avoid predicting the biased class, the model can more easily recognize other emotions for that group.


\section{Conclusion}\label{sec:conclusion}

    In this work, we investigate the impact of different types of dataset gender bias on how bias propagates into a model's predictions in the context of FER. To achieve this, we create a carefully designed, gender-balanced subset of AffectNet, a large ITW FER database. We induce two types of gender bias (representational and stereotypical) at varying strengths and configurations. Subsequently, we train a ResNet50 model on these biased datasets, evaluating its performance on a common test set to determine whether the model exhibits gender bias based on recall differences for each emotion class.


    Our results show that the impact of gender representational bias is less substantial than expected. Models exhibit a reasonable generalization ability even in the absence of one gender from the training dataset. In contrast, single-class stereotypical bias has a more pronounced effect despite involving a smaller perturbation in the dataset. With stereotypical bias, our results show a stronger effect specifically on the biased class, with additional impact observed on non-biased classes (especially when biasing the largest and better recognizable classes). While these results align with some previous works~\cite{Dominguez-Catena2023}, they diverge from other that observed larger effects stemming from representational bias~\cite{Kim2021,Buolamwini2018}. This difference may be specific to the FER domain and to our experimental setup. Thus, it is still important not to overlook the potential impact of representational bias in other contexts. 

    As future lines of work, we aim to extend our investigation of bias propagation across various setups, models, problems, and datasets. Further research is warranted to understand the different behaviors exhibited by classes within the dataset, as the underlying reasons for these variations remain unclear. Similarly, our findings regarding the limited propagation of bias to other classes necessitate deeper investigation. This includes complex multi-class stereotypical bias scenarios, which could reveal stronger interaction effects. Finally, the observed reduced impact of representational bias compared to stereotypical bias suggests promising research directions for mitigation strategies focused specifically on balancing the most severely biased classes, rather than the entire dataset.


    \section{Acknowledgments}

    This work was funded by a predoctoral fellowship from the Research Service of the Universidad Publica de Navarra, the Spanish MICIN ( PID2020-118014RB-I00 and PID2022-136627NB-I00/AEI/10.13039/501100011033 FEDER, UE), and the Government of Navarre (0011-1411-2020-000079 - Emotional Films).

\FloatBarrier
\bibliography{months,fulllibrary}

\clearpage
\appendix

\section{Detailed stereotypical bias results}\label{app:gender_disagg}

This Appendix contains additional figures illustrating the bias transference in the case of stereotypical bias for the remaining 6 emotions not included in Section~\ref{ssec:res_stereo}, namely, \textit{disgust}, \textit{fear}, \textit{sad}, \textit{surprise}, \textit{happy}, and \textit{neutral}. For each of the emotions, the corresponding figure (Figures~\ref{fig:recalldiffs_disgust} to~\ref{fig:recalldiffs_neutral}) includes the difference in recalls (female to male), the female recalls, and the male recalls, across the different values of the bias factor $f$.

    \def\emotions{
    disgust,
    fear,
    sad,
    surprise,
    happy,
    neutral}
    
    \foreach \emotion in \emotions
    {\begin{figure}[htbp]
        \centering
        \subfloat[\textbf{Difference in recalls} (F-M) under stereotypical bias ({\emotion} class)]{\includegraphics[width=\columnwidth]{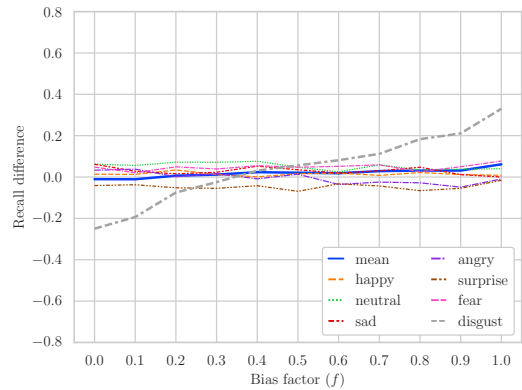}\label{sfig:recalldiffs_\emotion}}
        \vspace{-.7em}
        \subfloat[\textbf{Recall for the female group} under stereotypical bias ({\emotion} class)]{\includegraphics[width=\columnwidth]{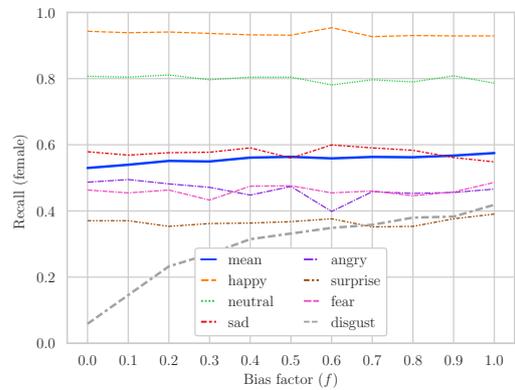}\label{sfig:recallfem_\emotion}}
        \vspace{-.7em}
        \subfloat[\textbf{Recall for the male group} under stereotypical bias ({\emotion} class)]{\includegraphics[width=\columnwidth]{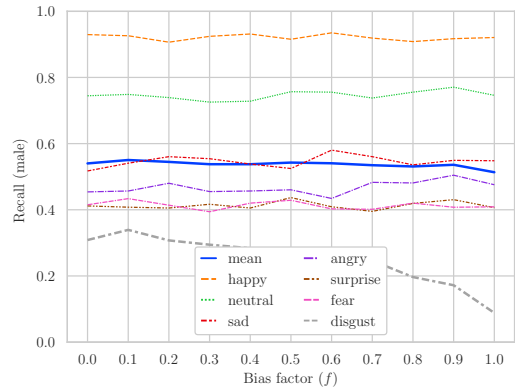}\label{sfig:recallmale_\emotion}}

        \caption{(a) Recall difference (female recall minus male recall) for each stereotypically biased dataset with biased class \textit{\emotion}. (b) Recall per class for the female group. (c) Recall per class for the male group. For all three, in the horizontal axis, amount of induced bias (bias factor $f$).}
        \label{fig:recalldiffs_\emotion}
    \end{figure}}

\end{document}